\documentclass[10pt,twocolumn,letterpaper]{article}

\usepackage[pagenumbers]{cvpr}

\definecolor{cvprblue}{rgb}{0.21,0.49,0.74}
\usepackage[pagebackref,breaklinks,colorlinks,allcolors=cvprblue]{hyperref}

\title{OrdiHand-GS: Ordinal-Guided Tracking and Hand-Refined Gaussian Avatars for Digital Signers}

\author{Mingxuan Ma \qquad Ronglai Zuo\\
Imperial College London\\
{\tt\small \href{https://mmxmichelle.github.io/OrdiHand-GS/}{https://mmxmichelle.github.io/OrdiHand-GS/}}}

\begin{document}
\twocolumn[{
\renewcommand\twocolumn[1][]{#1}
\maketitle
\begin{center}
\includegraphics[width=\textwidth]{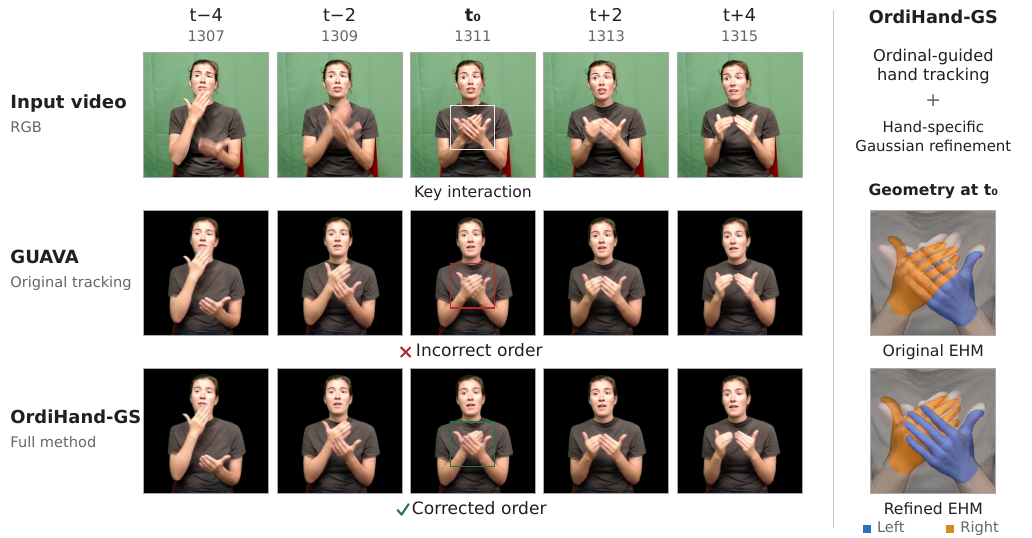}
\captionsetup{hypcap=false}\captionof{figure}{\textbf{OrdiHand-GS improves hand interactions in digital signers.} Five evenly spaced frames of a monocular signing clip, rendered from the same target frames and crop by GUAVA with its original tracking and by the complete OrdiHand-GS. At the central interaction frame, the original expressive-human tracking places the overlapping hands in the wrong front--back order; ordinal-guided refinement corrects the order, and the full method adds hand-specific Gaussian refinement. The inset shows the original and refined fits at that frame in the same view (blue: left hand, orange: right hand); its human annotation is left hand in front. We instantiate the avatar backbone with GUAVA.}\label{fig:teaser}
\end{center}
}]
\begin{abstract}
Digital signers, also known as signing avatars, convey sign language through animated virtual humans, where visual fidelity and linguistic intelligibility hinge primarily on the hands. When one hand crosses over the other, two failure modes dominate. First, monocular expressive-human tracking suffers from ordinal depth ambiguity: either hand can satisfy the same 2D landmarks, and the baseline tracker orders wrists at chance level on 365 annotated overlapping frames from How2Sign. Second, hand appearance remains blurred in neural rendering even after geometry is corrected. We propose OrdiHand-GS to address these errors separately. Relative depth from a video depth estimator and apparent hand size vote, with confidence weighting and abstention, for a front--back relation, which enters expressive-human tracking as a hinge on median hand-surface depth. A source-conditioned residual head then refines the local positions of hand-associated UV Gaussians with the avatar backbone and renderer frozen. We instantiate the framework with GUAVA. Against matched unconstrained refinement, wrist-order accuracy rises from 32.9\% to 77.3\% at a cost of 0.39 pixels of reprojection error, while a random-sign placebo gives no balanced-accuracy gain. A $2\times2$ factorial evaluation attributes most of the interaction-rendering improvement to tracking-stage refinement, with a smaller additional benefit from the residual. 
\end{abstract}

\section{Introduction}
Digital signers present sign language through an animated virtual human~\cite{forte2023reconstructing,baltatzis2024neural,yu2024signavatars,ivashechkin2025}. Gaussian-splatting avatars~\cite{kerbl2023,lei2024,shao2024,hu2024a,hu2024b,kocabas2024} make such signers drivable by tracked body, face and hand parameters and, most recently, reconstructable from a single source image~\cite{zhang2025}. For a digital signer the hands are the message: handshape, placement and the relative position of the two hands carry lexical and grammatical content~\cite{brentari1998}, so errors at the hands are salient to signing viewers~\cite{forte2023reconstructing,ivashechkin2025}. The pipeline has two stages that can each fail at the hands. The expressive human model (EHM) tracker~\cite{zhang2025,pavlakos2019} decides where the hands are and which one is nearer; the avatar decides how they look.

As a signer brings one hand across the other, the projected outlines of the two hands can agree with the video even if the reconstructed wrists pass on the wrong side of each other, because two mirror-image depth orders explain the same landmarks almost equally well, a known ambiguity of monocular hand and body reconstruction~\cite{pavlakos2018ordinal,zhang2021interacting,rong2021monocular}. Whichever order the optimizer settles on is passed to the avatar as an articulated surface whose Gaussians are splatted in that order, so the tracker, not the appearance model, sets the rendered occlusion, and a full-frame image score can miss the error because the interaction occupies a small part of the image. We quantify this failure on How2Sign~\cite{duarte2021}: on 365 overlapping-hand frames from 171 clips, an annotator viewing only RGB places the left hand in front in 252 cases, whereas the original EHM fit places the right wrist in front in 93.2\% of frames. Its balanced accuracy on wrist order is 49.8\%, and running the same objective for more iterations changes only four predictions. The objective lacks the evidence to break the tie.

Resolving this ambiguity raises two questions: what image evidence indicates front--back hand order, and at what level it should enter the optimization. Video Depth Anything (VDA)~\cite{chen2025video} produces relative depth that is consistent within a frame, which suffices to compare two hands even though it is not metric; on Re:InterHand~\cite{moon2023dataset}, where official 3D geometry defines the true order, visible-region VDA depth orders finger and hand regions correctly when confident but abstains often. Apparent size supplies a complementary cue at the scale of a whole hand, since the nearer hand tends to cover more pixels, although articulation and foreshortening can mislead it. Neither cue is reliable alone and each can be undecided, so both vote with a deterministic confidence and the votes are fused into one sign per frame.

Where the sign is applied matters as much as where it comes from. Constraining local pairs of hand parts lets a margin be satisfied by bending one finger while the wrists stay where they were. We instead impose a hinge on the median view-space depth of each hand's MANO-associated surface~\cite{romero2017} while resuming the EHM fit with its own optimizer. Gradients propagate through the SMPL-X parameterization~\cite{pavlakos2019}, and the evaluated optimizer retains the baseline pose, camera and shared-identity parameter groups, so its effect is a tracking-stage refinement rather than an isolated hand-pose correction. Against a matched control that refines for the same budget without the ordinal term, wrist-order accuracy improves from 32.9\% to 77.3\%, correcting 184 control errors and introducing 22, whereas a placebo that randomizes the signs produces no balanced-accuracy gain. The cost is 0.39 pixels of hand reprojection error and about 1.6 degrees of arm-chain rotation.

Correct ordering is necessary but not sufficient for convincing hands. Rendering the refined fits through an unchanged avatar backbone lowers perceptual error on the hand-interaction crop without changing the full-frame score, yet interaction LPIPS~\cite{zhang2018} stays near 0.31. Our second component is a lightweight, source-conditioned residual head that predicts a position offset for each UV Gaussian and applies it through a continuous hand-ownership weight, so that only hand-associated Gaussians move while the original feature extractors, decoders and neural refiner stay frozen. Because the head is trained on original tracking and the same checkpoint is then applied unchanged to refined tracking, the two components can be evaluated in a $2\times2$ factorial design on the same held-out frames (Fig.~\ref{fig:teaser}). This design attributes most of the interaction-rendering gain to the tracking stage, finds a smaller additional gain from the residual, and does not support a synergistic interaction between the two.

In summary, our main contributions are:
\begin{itemize}
    \item \textbf{Ordinal-guided EHM refinement.} We fuse VDA relative depth and apparent hand size by a confidence-weighted vote with abstention, and impose the resulting sign through a hand-level depth hinge while resuming expressive-human tracking. The constraint has no learned parameters and raises wrist-order accuracy on overlapping signing frames from 32.9\% to 77.3\% over a budget-matched control.
    \item \textbf{Hand-specific UV Gaussian residuals.} We introduce an ownership-gated, source-conditioned residual head that refines the local positions of hand-associated UV Gaussians with the avatar backbone and renderer frozen, and show that it improves hand rendering on both original and refined tracking.
    \item \textbf{A controlled evaluation of hand ordering and rendering in digital signers.} We construct 365 human-annotated overlapping-hand frames from 171 How2Sign clips and evaluate with a matched-budget control, a random-sign placebo, overlap stratification and a $2\times2$ factorial rendering study, which show that the tracking stage supplies most of the rendering improvement.
\end{itemize}

\section{Related Work}
\noindent\textbf{Gaussian avatars and digital signers.}
Neural human avatars first used implicit radiance fields~\cite{mildenhall2021,peng2021,su2021,chen2021,chen2023,zhao2022,yu2023}; Gaussian human-avatar methods~\cite{kerbl2023,hu2024b,kocabas2024,lei2024} combine explicit rendering primitives with articulated templates, and head-avatar methods rig Gaussians to a tracked face mesh~\cite{qian2024}. Per-subject monocular methods fit an avatar to the target person's video: GaussianAvatar~\cite{hu2024gaussianavatar} predicts pose-dependent Gaussians on a body model, SplattingAvatar~\cite{shao2024} binds Gaussians to a deformable mesh, and ExAvatar~\cite{moon2024expressive} and EVA~\cite{hu2024a} drive SMPL-X Gaussians with animatable hands and face. GUAVA~\cite{zhang2025} reconstructs a generalizable upper-body avatar from one source image and is our backbone; none of these methods constrains the front--back order of the two hands (Sec.~\ref{sec:main_results}). Signing avatars have been reconstructed from video with linguistic priors~\cite{forte2023reconstructing}, driven from text or gloss by sign language production models~\cite{zelinka2020,saunders2022,huang2022,fang2025,zuo2025,baltatzis2024neural,yu2024signavatars,zuo2026madis,zuo2024simple}, and rendered with Gaussians by SignSplat~\cite{ivashechkin2025}, which addresses temporally consistent fitting and Gaussian regularization; text-to-motion generators~\cite{zhang2023,jiang2023}, their whole-body extensions~\cite{lu2024}, and video-based human synthesis~\cite{chan2018,liu2019} generate motion or appearance without constraining the depth order of the hands. Signing avatars address deaf communities for whom signed languages are primary languages~\cite{senghas2002}, and deaf viewers judge such avatars by the naturalness of their movement and the readability of hand shapes and facial expression~\cite{kipp2011assessing}, so hand fidelity is a central requirement. We focus on the ordinal ambiguity of overlapping hands and distinguish changes in tracking from an additional, localized rendering adaptation.

\noindent\textbf{Expressive-human and hand reconstruction.}
Parametric body models go back to SCAPE~\cite{anguelov2023}, which early work fitted to single images from sparse keypoints and silhouettes~\cite{guan2009,chen2013,cheng2016} or used to reshape bodies in photographs~\cite{zhou2010}. SMPL~\cite{loper2015} and MANO~\cite{romero2017} model articulated bodies and hands, SMPL-X~\cite{pavlakos2019} combines body, hands and face~\cite{li2017}, and whole-body and hand regressors estimate detailed hands from images~\cite{moon2022,pavlakos2024reconstructing}. Interacting-hand datasets and methods~\cite{moon2020interhand2,zhang2021interacting,li2022interacting,rong2021monocular} show that close interactions remain ambiguous under monocular projection: similar 2D alignments can correspond to different front--back relations. We resume an existing EHM fit and constrain a hand-surface depth statistic, rather than independently reconstructing two hands.

\noindent\textbf{Depth estimation and ordinal evidence.}
Relative depth is useful when the desired observation is which surface lies closer, rather than its metric distance~\cite{zoran2015learning,chen2016single,xian2018monocular}, and ordinal depth has also supervised human pose estimation~\cite{pavlakos2018ordinal}. We use frozen Video Depth Anything predictions~\cite{chen2025video}, which build on affine-invariant monocular depth models~\cite{ranftl2020towards,yang2024depth}, as ordinal evidence. Apparent area supplies a complementary but articulation-dependent cue. Unlike directly matching predicted metric depth, our constraint uses a sign and a deterministic confidence weight.

\begin{figure*}[t]\centering
\includegraphics[width=\textwidth]{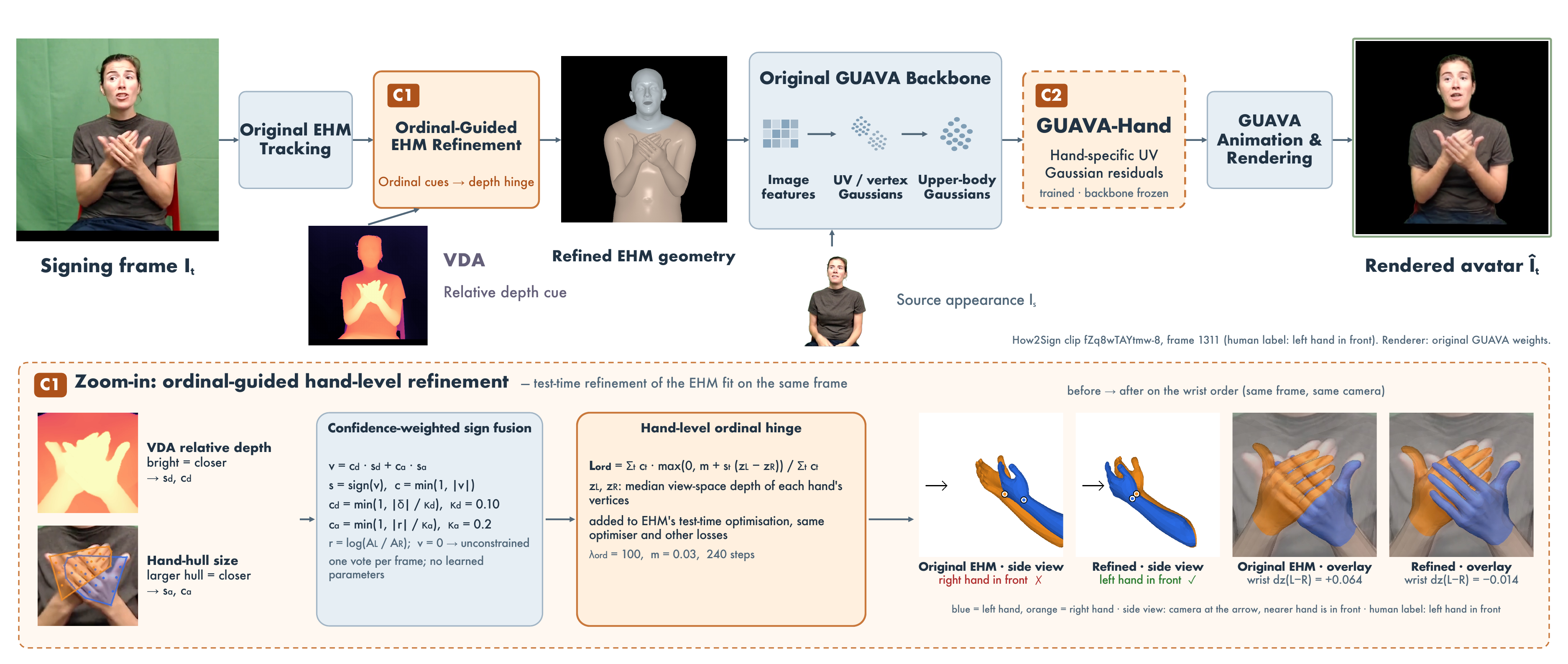}
\caption{\textbf{OrdiHand-GS framework.} Ordinal cues from VDA relative depth and projected hand-hull size are fused into a confidence-weighted sign and imposed through a hand-level depth hinge during EHM refinement (C1; zoom-in below). The refined EHM geometry drives the frozen GUAVA backbone with the source appearance $I_s$; $I_t$ is the input frame and $\hat I_t$ the rendered avatar. The dashed GUAVA-Hand block (C2) is the ownership-gated UV position residual of Sec.~\ref{sec:hand_refinement}; it is trained with the backbone and renderer frozen and evaluated with C1 in the factorial study. The zoom-in shows How2Sign clip fZq8wTAYtmw-8, frame 1311 (label: left hand in front), where refinement moves the left wrist (blue) in front of the right (orange); dz is the left-minus-right wrist-joint depth, negative when the left hand is nearer, and in the fusion box $s$ and $c$ are a cue's sign and confidence, subscripts $d$ and $a$ mark the depth and size cues, and the per-frame subscript $t$ of Sec.~\ref{sec:ordinal} is omitted.}\label{fig:method}
\end{figure*}

\section{Method}
\label{sec:method}

OrdiHand-GS addresses two sources of error in digital signers: ambiguous front--back hand ordering in monocular tracking, and residual hand-appearance error in avatar rendering. We first add image-derived ordinal constraints to expressive human model (EHM) fitting, then add a lightweight hand-specific position residual to a source-conditioned Gaussian avatar. We instantiate the avatar backbone with GUAVA~\cite{zhang2025}, which reconstructs a generalizable upper-body avatar from a single source image; the two components are optimized separately (Fig.~\ref{fig:method}).

\subsection{Overview and Parameterization}
\label{sec:base}
Given a monocular signing video, the EHM tracker supplies per-frame articulated parameters and camera extrinsics, together with shared identity parameters. Our refinement resumes these estimates using the baseline EHM optimizer. In the evaluated implementation, its parameter groups include body and hand poses, camera rotation and translation, and shared shape, scale and joint-offset parameters, which receive a nonzero learning rate only in the first optimization batch. We retain the baseline fitting and regularization terms and add an ordinal penalty.

Avatar reconstruction uses a source image $I_s$ and its EHM fit to predict vertex- and UV-associated Gaussians. Target EHM parameters deform these Gaussians for rendering. Our learned residual modifies only the local positions of hand-associated UV Gaussians. The original feature extractors~\cite{oquab2023}, Gaussian decoders and neural image refiner remain frozen during residual training.

\subsection{Ordinal-Guided EHM Refinement}
\label{sec:ordinal}
\noindent\textbf{Image-space evidence.}
A frozen Video Depth Anything model~\cite{chen2025video} provides a relative-depth score $d_t$, with larger values indicating closer surfaces. We build a dilated convex hull $M_{h,t}$ from the detected keypoints~\cite{yang2023effective} of each hand $h\in\{L,R\}$. Removing a dilated intersection of the two hulls gives non-overlapping support regions $S_{L,t}$ and $S_{R,t}$. We compute
\begin{equation}
\delta_t = \frac{\operatorname{median}_{S_{L,t}}d_t-
\operatorname{median}_{S_{R,t}}d_t}
{P_{95}(d_t|_{S_{L,t}\cup S_{R,t}})-P_5(d_t|_{S_{L,t}\cup S_{R,t}})}.
\end{equation}
The depth cue abstains when either support is too small, the denominator is near zero, or $|\delta_t|$ is below a threshold. Otherwise the cue votes with sign $s_{d,t}=\operatorname{sign}(\delta_t)$ and confidence $c_{d,t}=\min(1,|\delta_t|/\kappa_d)$, where $\kappa_d$ is the value of $|\delta_t|$ at which confidence saturates. Throughout, a positive ordinal sign denotes the signer's left hand in front.

As a complementary heuristic, we use the apparent sizes of the two hands. Let $A_{h,t}=|M_{h,t}|$ be the area of the dilated hull before overlap removal, and let $r_t=\log(A_{L,t}/A_{R,t})$. Away from near ties, the size cue votes with sign $s_{a,t}=\operatorname{sign}(r_t)$ and confidence $c_{a,t}=\min(1,|r_t|/\kappa_a)$, with saturation constant $\kappa_a$. This cue assumes that a larger projected hand tends to be closer, but is also affected by articulation and foreshortening. An abstaining cue has zero sign and confidence. We fuse the two cues as
\begin{align}
 v_t&=c_{d,t}\,s_{d,t}+c_{a,t}\,s_{a,t},\notag\\
 s_t&=\operatorname{sign}(v_t),\qquad c_t=\min(1,|v_t|).
\end{align}
These confidence weights are deterministic cue strengths, not calibrated probabilities. Conflicting cues reduce the resulting weight; an exactly cancelled vote receives no ordinal penalty.

\noindent\textbf{Hand-level constraint.}
Let $z_{h,t}$ be the median view-space depth of the MANO-associated EHM vertices of hand $h$, with increasing $z$ pointing away from the camera. For the frames in an optimization batch, we use
\begin{align}
 \mathcal{L}_{\mathrm{ord}}&=
 \frac{\sum_t c_t\,[m+s_t(z_{L,t}-z_{R,t})]_+}
 {\sum_t c_t+\epsilon},\\
 \mathcal{L}_{\mathrm{fit}}&=\mathcal{L}_{\mathrm{EHM}}
 +\lambda_{\mathrm{ord}}\mathcal{L}_{\mathrm{ord}}.
\end{align}
Here $m$ is the depth margin, $[\cdot]_+=\max(0,\cdot)$, $\epsilon$ is a small constant that keeps the normalization defined when all cues abstain, and $\lambda_{\mathrm{ord}}$ weights the ordinal term, which is zero when all cues abstain. It operates on a hand-surface statistic, with gradients propagating through the EHM parameterization, rather than imposing a rigid translation on an isolated hand. Satisfying the median-depth margin does not guarantee correct ordering at every finger or resolve contact and penetration. Evaluation reads out wrist order separately from this surface-depth constraint.

We use $m=0.03$ in EHM coordinate units and $\lambda_{\mathrm{ord}}=100$. Refinement takes 240 steps with hand-pose learning rate $10^{-4}$; other parameter groups retain the baseline settings. The main method uses unsmoothed fused votes. Frames with $c_t=0$ receive no ordinal penalty but remain affected by the baseline terms; a temporally smoothed cue is evaluated only as an ablation.

\subsection{Hand-Specific UV Gaussian Residuals}
\label{sec:hand_refinement}
The source reconstruction network predicts a frozen UV feature map $F$ and baseline local positions $\mathbf{p}_i$ for mesh-bound UV Gaussians. A convolutional head $f_\phi$ predicts a three-dimensional residual at each UV location. For Gaussian $i$ bound to a triangle with vertices $v_{ik}$ and barycentric weights $b_{ik}$, we define continuous hand ownership
\begin{equation}
 w_i=\sum_{k=1}^3 b_{ik}h(v_{ik}),\qquad
 \mathbf{p}'_i=\mathbf{p}_i+w_i[f_\phi(F)]_i,
\end{equation}
where $h(v)$ is one for MANO-associated hand vertices and zero elsewhere. Residuals vanish outside hand-associated faces and are blended at boundary faces. The corrected position is expressed in the binding triangle's local frame, so the residual is applied before splatting and remains attached to the articulated mesh.

The three-layer convolutional head has 221,571 trainable parameters, its final layer is zero-initialized so that training starts from the baseline output, and it predicts only UV local-position residuals; the vertex-Gaussian branch and the baseline color, opacity, rotation and scale predictions are unchanged.

\subsection{Optimization and Inference}
\label{sec:training_inference}
We train the residual head on source--target image pairs with the original tracking, retaining the backbone image objective and adding hand-crop SSIM supervision. Let $\mathcal{Y}=\{\hat I^{\mathrm{raw}},\hat I\}$ contain the raw splatting RGB and the final neural-refined RGB, let $I$ be the target frame, and let $C_h$ and $C_{\mathrm{head}}$ denote fixed crop operators. Define
\begin{align}
 \ell_{\mathrm{img}}(X,I)&=\|X-I\|_1+0.05\,\mathrm{LPIPS}_{\mathrm{Alex}}(X,I),\\
 \ell_{\mathrm{hand}}(X,I)&=0.1\|X-I\|_1
 +0.005\,\mathrm{LPIPS}_{\mathrm{Alex}}(X,I)\notag\\
 &\quad+0.01\,[1-\mathrm{SSIM}(X,I)].
\end{align}
The image-dependent objective is
\begin{align}
 \mathcal{L}_{\mathrm{render}}=\sum_{X\in\mathcal{Y}}\Big[&\ell_{\mathrm{img}}(X,I)
 +0.25\,\ell_{\mathrm{img}}(C_{\mathrm{head}}X,C_{\mathrm{head}}I)\notag\\
 &+\sum_{h\in\{L,R\}}\ell_{\mathrm{hand}}(C_hX,C_hI)\Big].
\end{align}
Ground-truth backgrounds are masked to black and training crops are resized to $256\times256$. The existing local-position and scale regularizers are also computed, but they depend only on frozen baseline predictions and do not regularize the added residual. Gradients from image losses pass through the frozen renderer to $\phi$.

The formal run uses a fixed final checkpoint after 40,000 residual-only updates. At inference, ordinal refinement supplies the target EHM parameters, while source reconstruction supplies the avatar and its learned hand residuals. The same residual checkpoint is applied to original and fused tracking without retraining, enabling a $2\times2$ evaluation of tracking-stage refinement and Gaussian residual refinement. Training settings and cue thresholds are given in Sec.~\ref{sec:experiments} and the supplementary material.

\section{Experiments}
\label{sec:experiments}

\begin{table}[t]\centering\footnotesize\setlength{\tabcolsep}{2pt}
\caption{Comparison with monocular avatar methods on 127 annotated frames from 60 clips. Per-subject methods are fitted to 140 frames of each clip; GUAVA-based rows use one reference image. $\dagger$: SMPL-X adapted, stage~1. $\ddagger$: SMPL-X mesh.}\label{tab:methods}
\begin{tabular}{lrrrrr}\toprule
&Full&\multicolumn{3}{c}{Interaction crop}&Hand\\\cmidrule(lr){3-5}
Method&LPIPS $\downarrow$&LPIPS $\downarrow$&SSIM $\uparrow$&PSNR $\uparrow$&LPIPS $\downarrow$\\\midrule
EVA~\cite{hu2024a}&\textbf{0.0940}&0.3237&0.6942&16.51&0.3467\\
ExAvatar~\cite{moon2024expressive}&0.1145&0.3510&0.6873&16.08&0.3793\\
GaussianAvatar$^\dagger$~\cite{hu2024gaussianavatar}&0.1058&0.3435&0.6686&15.26&0.3641\\
SplattingAvatar$^\ddagger$~\cite{shao2024}&0.1252&0.3627&0.6825&\underline{17.35}&0.3822\\
GUAVA~\cite{zhang2025}&0.0986&\underline{0.3190}&\underline{0.7064}&16.95&\underline{0.3454}\\
OrdiHand-GS (ours)&\underline{0.0977}&\textbf{0.3026}&\textbf{0.7209}&\textbf{17.60}&\textbf{0.3236}\\\bottomrule
\end{tabular}
\end{table}

\begin{table*}[t]\centering\small
\caption{Wrist order on 365 How2Sign frames (clip-clustered intervals; accuracy changes in percentage points; corrections and new errors relative to matched control). Rows above the rule are the main comparison, the rest the ablations of Sec.~\ref{sec:ablation}.}\label{tab:order}
\begin{tabular}{lrrrrr}\toprule Tracking arm & Accuracy [95\% CI] & BA [95\% CI] & $\Delta$accuracy vs.\ control [95\% CI] & Corrected & New errors\\\midrule
Original EHM &33.4 [28.1,39.1]&49.8 [46.4,52.7]&---&---&---\\
Control &32.9 [27.5,38.3]&48.9 [45.5,52.0]&0&---&---\\
\textbf{Fused (proposed)} &77.3 [72.1,81.5]&76.9 [71.5,81.7]&+44.4 [36.9,51.6]&184&22\\\midrule
Local VDA hook &44.4 [39.1,49.2]&56.5 [52.4,60.2]&+11.5 [7.7,15.5]&52&10\\
VDA hinge &67.1 [62.2,71.7]&67.4 [61.9,72.5]&+34.2 [27.0,41.3]&156&31\\
Local hook and VDA hinge &71.0 [65.9,75.7]&70.7 [65.3,75.7]&+38.1 [30.9,45.6]&170&31\\
Area hinge &73.7 [68.4,78.4]&74.9 [69.5,79.9]&+40.8 [33.6,47.9]&174&25\\
Random-sign hinge &42.2 [37.2,47.6]&48.6 [43.5,53.7]&+9.3 [3.7,15.0]&71&37\\
Overlap-gated &77.0 [72.5,81.2]&78.7 [73.9,83.1]&+44.1 [37.3,50.4]&179&18\\
Fused, smoothed &77.8 [72.9,82.0]&77.6 [72.3,82.1]&+44.9 [37.9,51.8]&185&21\\\bottomrule\end{tabular}
\end{table*}
\noindent\textbf{Datasets and annotation.}
We evaluate on the How2Sign~\cite{duarte2021} test set (184 clips). From frames whose DWPose~\cite{yang2023effective} hand hulls overlap by at least 0.30, one annotator labeled the front hand from RGB triplets without seeing model outputs, giving 365 determinate labels from 171 clips (252 left-front, 113 right-front; hidden-repeat $\kappa=0.91$). Labels are used only for evaluation. The local-cue check uses 40 Re:InterHand~\cite{moon2023dataset} frames, where official 3D geometry defines the order of 665 finger--hand and finger--finger relations.

\noindent\textbf{Metrics.}
Tracking is scored by wrist-order accuracy and balanced accuracy (BA, the mean of the two class recalls), with corrections and new errors counted relative to control. Rendering is scored by VGG LPIPS~\cite{zhang2018}, SSIM~\cite{wang2004}, PSNR and L1 on a fixed square interaction crop enclosing both hand hulls, on per-hand crops, and on the full frame as a safety check; crops are derived from the original keypoints and shared across arms. All intervals are 95\% clip-clustered bootstrap intervals, and arm differences are paired by frame.

\noindent\textbf{Compared arms.}
Every tracking arm resumes the same original EHM fit with the same optimizer, seed, frames and budget. \emph{Control} sets the ordinal weight to zero and \emph{fused} is the proposed confidence-weighted vote; the ablation arms are defined in Sec.~\ref{sec:ablation}. Rendering follows a $2\times2$ design: C0 (original tracking, original weights), C1 (fused tracking, original weights), C2 (original tracking, learned residual) and C3 (fused tracking, the same residual checkpoint). All arms share the source image and source-frame pose record, while shared identity follows each tracking arm.

\noindent\textbf{Compared methods.}
Table~\ref{tab:methods} compares OrdiHand-GS and unmodified GUAVA~\cite{zhang2025} with four per-subject monocular methods, EVA~\cite{hu2024a}, ExAvatar~\cite{moon2024expressive}, GaussianAvatar~\cite{hu2024gaussianavatar} and SplattingAvatar~\cite{shao2024}, the latter two adapted to SMPL-X. This comparison uses a signer-stratified subset of 60 of the 171 labeled clips, drawn without reference to labels or model outputs, with all 127 annotated frames they contain. Each per-subject method is fitted with its official code and training length to 140 frames of the same clip, all at least 2.5\,s from every evaluation frame, and estimates evaluation poses with its own pipeline; GUAVA and OrdiHand-GS take one reference image under the same rule, with original and fused EHM tracking. Regions and crop boxes follow the protocol above; rendering details and code changes are listed in the supplementary material.

\noindent\textbf{Implementation details.}
Refinement uses the settings of Sec.~\ref{sec:ordinal}. Sampled nine-frame neighborhoods of each clip are merged before optimization, so the inherited motion term also links entries across temporal gaps (Sec.~\ref{sec:limitations}). The residual head is trained for 40,000 updates on a video-disjoint split of 100,002 frames with the backbone frozen; remaining hyperparameters are in the supplementary material.

\begin{table}[t]\centering\small\setlength{\tabcolsep}{5pt}
\caption{Factorial rendering evaluation on the same 365 frames. T: fused tracking; H: learned hand residual. Interaction-crop metrics are from the formal training run.}\label{tab:factorial}
\begin{tabular}{lccrrrr}\toprule
Arm&T&H&LPIPS $\downarrow$&SSIM $\uparrow$&PSNR $\uparrow$&L1 $\downarrow$\\\midrule
C0&--&--&0.3174&0.7055&16.91&0.0897\\
C1&yes&--&0.3065&0.7146&17.35&0.0865\\
C2&--&yes&0.3155&0.7092&17.07&0.0881\\
C3&yes&yes&\textbf{0.3047}&\textbf{0.7185}&\textbf{17.50}&\textbf{0.0851}\\\bottomrule
\end{tabular}
\end{table}

\begin{table*}[t]\centering\small
\caption{Paired rendering with the original GUAVA weights on the 365 annotated frames; hand LPIPS for anatomical left and right; dashes: not reported.}\label{tab:render}
\begin{tabular}{lrrrrrrr}\toprule &\multicolumn{4}{c}{Interaction crop}&\multicolumn{2}{c}{Hand LPIPS}&Full frame\\
Tracking & LPIPS $\downarrow$ & SSIM $\uparrow$ & PSNR $\uparrow$ & L1 $\downarrow$ & Left & Right & LPIPS\\\midrule
Original (C0) &0.3174&0.7055&16.91&0.0897&0.3385&0.3493&0.0982\\
Matched control &0.3133&0.7097&17.13&0.0877&0.3342&0.3449&0.0981\\
Area hinge &0.3072&0.7138&17.30&0.0867&---&---&---\\
VDA hinge &0.3084&0.7130&17.28&0.0870&---&---&---\\
Fused (C1) &0.3065&0.7146&17.35&0.0865&0.3247&0.3299&0.0977\\\bottomrule\end{tabular}
\end{table*}

\subsection{Main Results}
\label{sec:main_results}
\noindent\textbf{Comparison with existing methods.}
Table~\ref{tab:methods} compares OrdiHand-GS with its backbone GUAVA and four per-subject methods on the 127-frame subset. Although it uses one reference image where the per-subject methods are fitted to 140 frames per clip, OrdiHand-GS has the best value on every interaction and hand metric: interaction LPIPS 0.3026 (0.3190 for GUAVA and 0.3237 for EVA, the best per-subject method on this crop), interaction SSIM 0.7209, PSNR 17.60 and hand LPIPS 0.3236. In a paired comparison, the interaction and hand PSNR gains over SplattingAvatar are not significant (supplementary Sec.~G). On the full frame the per-subject methods lead: EVA has the lowest LPIPS and SplattingAvatar the highest PSNR (supplementary Table~S5), because per-subject fitting captures identity better than one image, although OrdiHand-GS still improves on GUAVA. This frame set differs from that of Tables~\ref{tab:factorial} and \ref{tab:render}.

\noindent\textbf{Wrist order.}
To locate where these hand gains originate, we next evaluate front--back wrist order on all 365 annotated frames (Table~\ref{tab:order}). Original EHM misorders 243 of 365 frames from 142 clips. Its median absolute wrist-depth difference is 0.044 model units at a camera distance of roughly 11, and its BA interval includes chance. The matched control leaves this unchanged. The proposed fused refinement reaches 77.3\% accuracy and 76.9\% BA, a gain of 44.4 accuracy points [36.9, 51.6] over control; it corrects 184 control errors and introduces 22.

\noindent\textbf{Rendering.}
Table~\ref{tab:factorial} compares the four arms of the $2\times2$ design on the interaction crop. The complete system (C3) outperforms the original pipeline (C0) on every interaction metric, reducing LPIPS by 0.0127 [0.0114, 0.0146]. Most of this gain comes from the tracking stage: fused tracking alone (C1) is better than both C0 and the matched control (Table~\ref{tab:render}) on LPIPS, SSIM and PSNR, with paired LPIPS and SSIM intervals that exclude zero. The learned residual adds a smaller but consistent improvement on either tracking (C2 over C0 and C3 over C1), again with paired LPIPS and SSIM intervals excluding zero. The two contributions are close to additive: the exploratory interaction contrast $(\mathrm{C3}-\mathrm{C2})-(\mathrm{C1}-\mathrm{C0})$ is $+0.0001$ [$-0.0003$, $+0.0003$] for LPIPS, so no synergy is claimed. Relative to control, interaction LPIPS decreases in both overlap strata, and full-frame metrics do not worsen, although C2 full-frame LPIPS worsens slightly under heavy overlap. These differences include the effects of the updated camera and shared identity.

\subsection{Ablation Study}
\label{sec:ablation}
The lower rows of Table~\ref{tab:order} vary one element of the constraint at a time under the same budget.

\noindent\textbf{Cue choice.}
With a single cue, the area hinge outperforms the VDA hinge (Table~\ref{tab:order}); fusing the two is best, and gating by hull overlap (VDA below 0.44, area otherwise) performs comparably to fusion. These differences are small relative to their intervals: on the 341 frames covered by both cues the VDA minus area BA difference includes zero, and the gain of fusion over area alone, 3.6 points [$-1.1$, 8.4], remains uncertain (Sec.~\ref{sec:analysis}). The same ordering holds after rendering (Table~\ref{tab:render}): either single-cue hinge already recovers most of the interaction-crop improvement of the fused arm over control.

\noindent\textbf{Constraint level.}
Applying the VDA sign to local part pairs within and across hands is much worse than the hand-level hinge driven by the same sign (Table~\ref{tab:order}), and adding the local hook to the hand-level hinge gives only a modest further gain. At the smaller 60-step budget the local hook gains 3.6 points over its matched control [1.4, 6.0]. The local hook also increases the number of intersecting hand vertices relative to control, whereas the hand-level constraint does not (Sec.~\ref{sec:analysis}).

\noindent\textbf{Random-sign placebo.}
Randomizing the signs while keeping the active frames and confidences gains 9.3 raw accuracy points, but its BA difference from control is $-0.3$ [$-5.6$, 4.8]. The raw gain reflects class imbalance: the control places the right wrist in front in most frames while 69\% of labels are left-front, so random signs that push predictions toward an even split gain on the majority class, lose on the minority class and leave balanced accuracy unchanged. It also increases intersecting vertices and wrist motion most (supplementary Table~S3); the sign information, not the act of constraining, drives balanced order recovery.

\noindent\textbf{Temporal smoothing.}
A temporal median over the votes raises the constrained centers from 334 to 355, with a 0.5-point accuracy change [$-1.6$, 2.7]. It also reduces mean wrist speed and acceleration (supplementary Table~S3).

\begin{table}[t]\centering\footnotesize
\caption{Cue readout on covered frames (top) and optimized wrist-order accuracy per overlap stratum (bottom).}\label{tab:strata}
\begin{tabular}{lrrr}\toprule Cue & Coverage & Accuracy & BA\\\midrule
VDA &95.6&81.4&76.9\\
Hull area &97.3&85.1&80.5\\
Always left &100.0&69.0&50.0\\\midrule
Overlap ($n$)&VDA hinge&Area hinge&Fused\\\midrule
Light (182)&74.2&70.3&78.6\\
Heavy (183)&60.1&77.0&76.0\\\bottomrule\end{tabular}
\end{table}

\subsection{Analysis}
\label{sec:analysis}
\begin{figure*}[t]\centering
\includegraphics[width=0.85\textwidth]{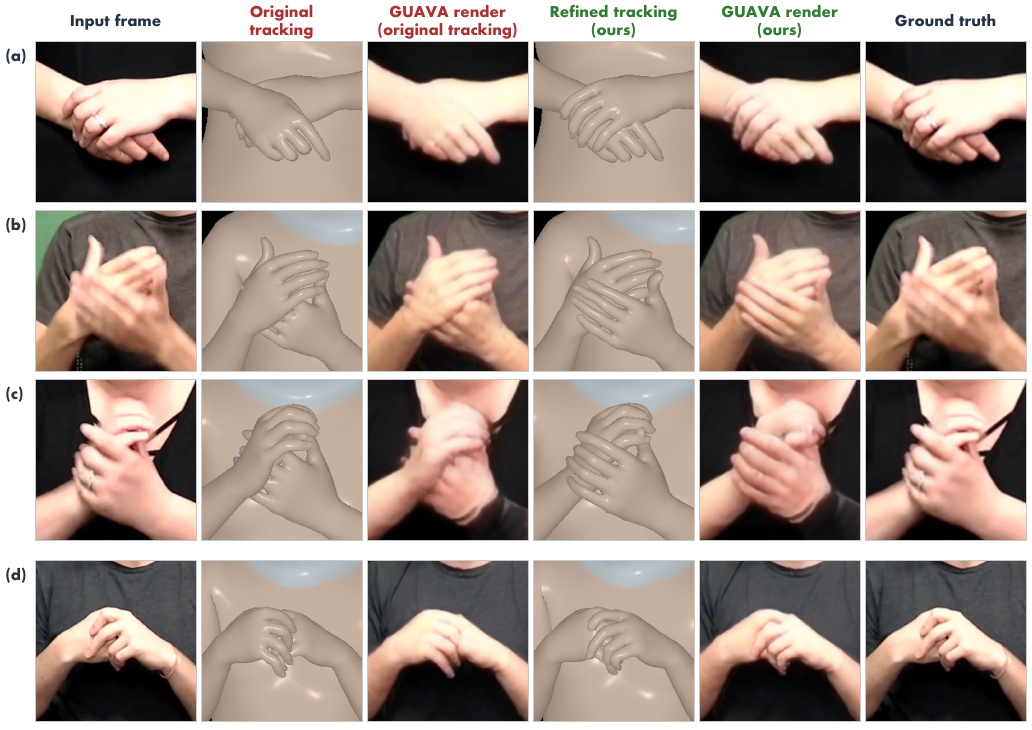}
\caption{\textbf{Ordinal refinement on interaction frames.} Columns: input, original EHM tracking and its GUAVA render, refined (fused) tracking and its render, ground truth; all renders use the original GUAVA weights (C0 vs.\ C1). Rows (a)--(c) are labeled left-front frames misordered by the original fit and corrected by refinement: (a) has the fourth-largest interaction-crop LPIPS gain of fused over control, (b)--(c) are random draws from the 184 corrected frames. Row (d) is unlabeled: the wrist order is flipped but the interlaced fingers are unchanged; finger-level ordering is future work.}\label{fig:crops}
\end{figure*}
\noindent\textbf{Validity of the local depth cue.}
On the Re:InterHand relations, visible-region VDA depth has the highest balanced accuracy but covers under a third of the relations, whereas apparent area is less accurate but covers almost all (supplementary Table~S1). The depth cue is therefore a selective vote; that analysis scores local surface pairs, not one wrist order per frame.

\noindent\textbf{Complementarity across overlap regimes.}
Table~\ref{tab:strata} scores each cue directly on the frames it covers and splits the optimized arms at the median hull overlap of 0.44. Under heavy overlap the area hinge is 16.9 accuracy points [9.8, 24.3] above the VDA hinge; under light overlap the gap is $-3.8$ [$-12.8$, 5.1] and not significant. Relative to the area hinge, fusion adds 8.2 points [1.7, 14.7] on light-overlap frames but only $-1.1$ [$-6.3$, 4.3] on heavy-overlap frames. The depth cue therefore helps when the hands overlap lightly, whereas apparent size carries the signal under heavy overlap.

\noindent\textbf{Costs in fit and short-window motion.}
Relative to control, fused refinement increases hand reprojection error by 0.39 pixels [0.32, 0.47] (4.95 to 5.34 pixels on a 1024-pixel crop, still below the original fit's 6.74) and arm-chain rotation by $1.59^\circ$ [1.43, 1.74]; the interval for the number of intersecting hand vertices spans zero, no finger joint exceeds the $120^\circ$ diagnostic threshold, and wrist speed and acceleration increase by 0.0005 and 0.0011 model units per frame and per frame squared (supplementary Table~S3). These diagnostics use adjacent frames in sampled windows and do not establish long-term consistency (Sec.~\ref{sec:limitations}).

\subsection{Qualitative Results}
\label{sec:qualitative}
Figure~\ref{fig:crops} shows three corrected interaction frames and one unlabeled case with corrected wrist order but unchanged finger interleaving. Of eight slow-motion windows, six show the correct hand in front without visible flicker; in w06 (supplementary Fig.~S1) fused tracking reverses the order incorrectly, in w05 motion blur complicates the image score, and in w03 the two cues cancel on frame 1310, which smoothing resolves. These are qualitative inspections, not a stability benchmark.

\section{Conclusion and Discussion}
\label{sec:limitations}
OrdiHand-GS combines ordinal-guided EHM refinement with hand-specific UV Gaussian residuals for digital signers. On annotated How2Sign frames, fused cues raise wrist-order accuracy from 32.9\% to 77.3\% over a matched control. Tracking-stage refinement drives most of the interaction-rendering gain; the residual head adds less.

\noindent\textbf{Limitations and future work.}
The median surface statistic neither forces rigid hand motion nor recovers contact, and camera and shared-identity updates also enter the refinement, so a pose-only mechanism is not isolated. The residual changes only UV local positions, is not bounded by the retained regularizers, and was evaluated together with the added hand SSIM term. The optimizer links consecutive entries of merged per-clip neighborhoods even across long gaps; all matched arms share this, a gap-aware paired rerun is not included, and short-window diagnostics do not establish full-video stability. Labels come from one observer, tracker prefiltering excludes some difficult interactions, and How2Sign has no 3D ground truth; silhouette IoU and hand--body penetration were not measured, and ordinal cues can introduce new ordering errors. Training and evaluation are video-disjoint rather than signer-disjoint within one backbone, and rendering is evaluated on labeled interaction frames only; cross-backbone transfer, sign intelligibility and perceptual quality for signing users remain open.

\noindent\textbf{LLM usage statement.}
The authors used OpenAI's ChatGPT and Codex to assist with manuscript drafting, translation, LaTeX formatting, figure assembly, and comparison of method descriptions with implementation code. Quantitative values were transcribed from existing experiment records. The authors are responsible for reviewing the manuscript, figures, citations, and reported results before submission.

\noindent\textbf{Acknowledgements.}
R. Zuo was supported by EPSRC Project GNOMON (EP/X011364/1) and Turing AI Fellowship (EP/Z534699/1). The authors acknowledge the use of resources provided by the Isambard-AI National AI Research Resource (AIRR). Isambard-AI~\cite{mcintosh2024isambard} is operated by the University of Bristol and is funded by the UK Government’s Department for Science, Innovation and Technology (DSIT) via UK Research and Innovation; and the Science and Technology Facilities Council [ST/AIRR/I-A-I/1023].
{
    \small
    \bibliographystyle{ieeenat_fullname}
    \bibliography{main}
}

\clearpage
\maketitlesupplementary
\appendix
\setcounter{table}{0}\renewcommand{\thetable}{S\arabic{table}}
\setcounter{figure}{0}\renewcommand{\thefigure}{S\arabic{figure}}

\section{Annotation Protocol}
\label{sec:supp_annotation}
\noindent\textbf{Frame pool.}
From the How2Sign test set of 184 clips, candidate frames require a DWPose~\cite{yang2023effective} hand-hull overlap of at least 0.30, at least 30 frames between samples and at most four frames per clip (seed 0). Selection uses only 2D geometry and involves no model output.

\noindent\textbf{Labeling.}
One annotator viewed RGB triplets at frame offsets $-5$, $0$, $+5$ without seeing any tracking or rendering output and labeled which hand is in front, or marked the frame as indeterminate. The 365 determinate labels from 171 clips comprise 252 left-front and 113 right-front frames. Agreement on hidden repeats was 52/60 overall and 49/51 on repeats with two determinate labels ($\kappa=0.91$). The protocol measures one observer's repeatability, not inter-observer agreement. Labels are used only for evaluation and never enter any optimization.

\noindent\textbf{Local-cue check on Re:InterHand.}
On 40 Re:InterHand~\cite{moon2023dataset} frames, official 3D geometry and camera parameters define the ground-truth front--back order of local relations. Of 741 candidate relations, 665 have a determinate order; the remaining 76 are ambiguous or near ties and are excluded. We evaluate finger--hand and finger--finger relations. For selective predictors, \emph{coverage} is the fraction of determinate relations that receive a prediction; accuracy and BA are computed on the covered subset.

\begin{table}[t]\centering\small
\caption{Local front--back relations on Re:InterHand. Accuracy and BA are percentages on each predictor's covered subset; coverage is the fraction of the 665 determinate relations that receive a prediction.}\label{tab:local}
\begin{tabular}{lrrr}\toprule Predictor & BA & Accuracy & Coverage (\%)\\\midrule
Majority &50.00&44.21&100.00\\
Area &65.25&66.83&92.48\\
No-depth geometry &51.17&57.30&97.89\\
Visible-region VDA &85.64&84.21&31.43\\\bottomrule\end{tabular}
\end{table}
Table~\ref{tab:local} reports each predictor on its covered subset. For visible-region VDA, BA is 84.04\% at 36.13\% coverage on finger--hand relations and 88.84\% at 27.79\% coverage on finger--finger relations. These estimates describe visible, confident samples; Macro-F1 and class recalls are not reported in this draft.

\section{Ordinal Cues and Thresholds}
\label{sec:supp_cues}
\noindent\textbf{Hand support regions.}
For each hand, the convex hull of its DWPose~\cite{yang2023effective} keypoints is dilated by 8 pixels. The intersection of the two hulls, dilated by 4 pixels, is removed from both to obtain the non-overlapping supports $S_{L,t}$ and $S_{R,t}$. Hull areas $A_{h,t}$ used by the size cue are measured before overlap removal.

\noindent\textbf{Depth cue.}
Relative depth comes from a frozen ViT-L Video Depth Anything model~\cite{chen2025video} with larger values indicating nearer surfaces. The cue abstains when either support has fewer than 50 pixels, when the normalizing range $P_{95}-P_{5}$ is near zero, or when $|\delta_t|<0.02$. Otherwise $s_{d,t}=\operatorname{sign}(\delta_t)$ and $c_{d,t}=\min(1,|\delta_t|/\kappa_d)$ with $\kappa_d=0.10$.

\noindent\textbf{Size cue.}
With $r_t=\log(A_{L,t}/A_{R,t})$, the cue abstains for $|r_t|\le 0.01$; otherwise $s_{a,t}=\operatorname{sign}(r_t)$ and $c_{a,t}=\min(1,|r_t|/\kappa_a)$ with $\kappa_a=0.2$.

\noindent\textbf{Fusion and hinge.}
An abstaining cue contributes zero sign and zero confidence. Votes are fused per frame as in the main text; an exactly cancelled vote leaves the frame without an ordinal penalty. A positive sign always denotes the signer's left hand in front. The hinge uses $m=0.03$ in EHM coordinate units and $\lambda_{\mathrm{ord}}=100$; $\epsilon$ is a small positive constant that keeps the normalization defined when all cues in a batch abstain. All thresholds and weights are fixed constants and were set before the corresponding arm was evaluated. The unsmoothed fused rule constrains 334 of the 365 labeled centers; the smoothed variant constrains 355.

\section{Tracking Arms and Budgets}
\label{sec:supp_arms}
\begin{table}[t]\centering\small
\caption{Tracking arms. Every arm resumes the same original EHM fit with the same optimizer, seed, frames and budget.}\label{tab:supp_arms}
\begin{tabular}{p{2.3cm}p{5.0cm}}\toprule
Arm & Definition\\\midrule
Control & Ordinal weight set to zero.\\
Local VDA hook & VDA sign applied to local part pairs within and across hands with a 1\,cm margin.\\
VDA hinge & VDA sign alone in the hand-level hinge.\\
Area hinge & Size sign alone in the hand-level hinge.\\
Local hook and VDA hinge & Both terms.\\
Random-sign hinge & Signs randomized; active frames and confidences retained.\\
Fused (proposed) & Confidence-weighted vote of both cues.\\
Overlap-gated & VDA sign when hull overlap is below the median of 0.44, size sign otherwise.\\
Fused, smoothed & Median of the nonzero fused votes within two frames on either side.\\\bottomrule
\end{tabular}
\end{table}

\noindent\textbf{Budgets.}
The primary budget takes 240 steps with hand-pose learning rate $10^{-4}$; the other parameter groups (body pose, camera rotation and translation, shared shape, scale and joint offsets) keep the baseline settings. A second, prespecified budget of 60 steps at $10^{-5}$ is reported for the local hook and its matched control. Optimization proceeds in nine-frame windows; the sampled neighborhoods of each clip are merged before optimization, so the inherited motion term also links consecutive entries across temporal gaps, as discussed in the main text.

\noindent\textbf{Order of specification.}
The hand-level and combined arms followed the budget comparison; the area and placebo arms were added next; the fused and gated rules were specified before the overlap stratification. The smoothing ablation followed the observation of a cancelled vote in a diagnostic clip (frame 1310 of the illustrated clip); its median excludes zero votes, a choice made after inspecting sign files without labels. Each arm's constants were fixed before its evaluation.

\section{Rendering Protocol}
\label{sec:supp_render}
\noindent\textbf{Renderer.}
All arms that use the original weights render with the same pretrained GUAVA checkpoint (iteration 160{,}000), the same self-reenactment settings and seed 10. Neither the Gaussian representation nor the neural image refiner is trained in the C0/C1 comparisons. All arms share the source image and source-frame pose record, while the shared identity parameters follow each tracking arm and also enter source reconstruction.

\noindent\textbf{Regions and metrics.}
The interaction crop is the square enclosing both hand hulls derived from the original keypoints, with 15\% padding and a 64-pixel minimum side. Interaction and single-hand crops are resized to $280\times280$; full-frame metrics are computed at $512\times512$ as a safety check. On each region we report VGG LPIPS, SSIM, PSNR and L1 (mean absolute pixel error). Crops are fixed from the original keypoints and shared across arms, so they do not depend on any arm's output.

\section{Residual-Head Training}
\label{sec:supp_residual}
The residual head is a three-layer convolutional network with 221{,}571 trainable parameters whose final layer is initialized to zero, so training starts from the baseline output. It predicts only UV local-position residuals; the vertex-Gaussian branch and the baseline color, opacity, rotation and scale predictions are unchanged, and the feature extractors, Gaussian decoders and neural refiner remain frozen. Training uses source--target pairs with the original tracking on a video-disjoint split of 100{,}002 frames for 40{,}000 residual-only updates; the formal run uses seed 10. Ground-truth backgrounds are masked to black and training crops are resized to $256\times256$. The training LPIPS term uses the AlexNet backbone, whereas evaluation LPIPS uses the VGG backbone at $280\times280$; the loss weights are those of the main text. The residual parameters are optimized with Adam ($\beta_1=0$, $\beta_2=0.99$) at batch size 6, with the learning rate decayed linearly from $10^{-3}$ to $2.5\times10^{-4}$ over the 40{,}000 updates. The same checkpoint is applied to original and fused tracking without retraining.

\section{Statistics and Diagnostics}
\label{sec:supp_stats}
All intervals are 95\% clip-clustered bootstrap intervals with 2{,}000 draws (seed 0); arm differences are paired by frame. Balanced accuracy is the mean of the two class recalls; corrections and new errors are counted relative to the matched control.

\begin{table}[t]\centering\small
\caption{Paired changes from matched control to fused on the 365 annotated frames. Rotation magnitudes are measured against original EHM. Speed and acceleration use model units per frame and per frame squared.}\label{tab:cost}
\begin{tabular}{lr}\toprule Measure & Difference [95\% CI]\\\midrule
Body 2D error (px)&+0.36 [0.28, 0.44]\\
Hand 2D error (px)&+0.39 [0.32, 0.47]\\
Arm-chain rotation ($^\circ$)&+1.59 [1.43, 1.74]\\
Finger rotation ($^\circ$)&$-0.23$ [$-0.25$, $-0.21$]\\
Intersecting vertices&$-2.7$ [$-18.0$, 11.7]\\
Maximum penetration depth&$-0.002$ [$-0.003$, $-0.001$]\\
Wrist speed&+0.0005 [0.0004, 0.0006]\\
Wrist acceleration&+0.0011 [0.0008, 0.0014]\\\bottomrule\end{tabular}
\end{table}

\noindent\textbf{Costs in fit and short-window motion.}
Table~\ref{tab:cost} lists the paired changes from matched control to fused. Hand reprojection error is measured on a 1024-pixel crop and increases from 4.95 to 5.34 pixels, still below the original fit's 6.74 pixels. No finger joint exceeds the $120^\circ$ diagnostic threshold for implausible articulation, and the interval for the number of intersecting hand vertices spans zero; by contrast, the local hook and the random-sign placebo increase that count by 44.0 [33.1, 55.4] and 34.7 [21.9, 47.6]. Short-window motion diagnostics use the 2{,}742 adjacent pairs and 2{,}238 triplets within the nine-frame windows. Fused wrist speed differs from original EHM by $-0.0001$ [$-0.0003$, 0.0001]; smoothing reduces the mean speed to 0.0111 and acceleration to 0.0104, from 0.0115 and 0.0112 for fused, and the random-sign arm has the largest values, 0.0152 and 0.0193. These diagnostics characterize adjacent frames within sampled windows and do not establish full-video stability.

\section{Comparison with Existing Methods}
\label{sec:supp_methods}
\noindent\textbf{Clip subset.}
The 171 labeled test clips come from six signers. Within each signer, clips were ordered by a hash value and the first clips were taken, giving 60 clips (6, 4, 13, 16, 14 and 7 per signer) without reference to labels or model outputs; they contain 127 of the 365 annotated frames (16, 6, 27, 34, 27 and 17 per signer; 6 clips contribute one frame, 41 two and 13 three). The remaining 111 clips were not run because of compute and time limits.

\noindent\textbf{Fitting frames and reference images.}
Every per-subject method is fitted to the same 140 frames per clip, sampled uniformly in time from frames at least 2.5\,s away from every evaluation frame of that clip. The EVA preprocessing drops frames in which both hands are not detected, leaving 7{,}874 of the 8{,}400 frames; GaussianAvatar and SplattingAvatar reuse the EVA preprocessing and therefore the same frames. GUAVA and OrdiHand-GS take one reference image per clip under the same 2.5\,s rule. Five of the 60 clips (11 evaluation frames) required a different reference image from the C0/C3 runs and were re-rendered; the other clips reuse the C0 and C3 renders.

\noindent\textbf{Pose sources and rendering.}
No baseline uses EHM tracking. EVA uses its official preprocessing (HaMeR hand estimates and SMPLify-X fitting); ExAvatar uses its official fitting (DECA, Hand4Whole and mmpose), with identity estimated on the 140 fitting frames and then frozen; GaussianAvatar and SplattingAvatar reuse the SMPL-X parameters fitted by the EVA pipeline, from which SplattingAvatar builds its mesh. Evaluation-frame poses are estimated by each method's own pipeline from the test image; no method optimizes against the test image. Per-subject methods render the $1280\times720$ frame, which is mapped to $512\times512$ by one fixed transform; GUAVA and OrdiHand-GS render $512\times512$ directly. Metrics are computed at $512\times512$ on a black background in the three regions of Sec.~\ref{sec:supp_render}, whose crop boxes come from the original EHM tracking and are shared by all methods; hand metrics average the left and right crops.

\begin{table}[t]\centering\small
\caption{Training settings of the compared methods and wall-clock time per clip on one A800 GPU.}\label{tab:supp_methods_train}
\begin{tabular}{p{2.4cm}p{3.1cm}r}\toprule
Method & Training & Time\\\midrule
EVA & official default, 2{,}000 iterations & 1.45\,h\\
ExAvatar & official 5 epochs (10{,}500 iterations) & 1.76\,h\\
GaussianAvatar & stage 1 only, 200 epochs & 34.9\,min\\
SplattingAvatar & official 30{,}000 iterations & 46\,min\\
GUAVA & 160k-step backbone, no per-clip training & 0\\
OrdiHand-GS & frozen 160k backbone, residual head trained for 40k updates (global iteration 200k), no per-clip training & 0\\\bottomrule
\end{tabular}
\end{table}

\noindent\textbf{Training settings and code changes.}
All methods use their official code and official training lengths (Table~\ref{tab:supp_methods_train}); no hyperparameter was tuned on results, and every checkpoint is the prespecified training endpoint. Times for EVA and ExAvatar include preprocessing; GaussianAvatar additionally needs the EVA preprocessing. The following changes were made and recorded before the runs. \emph{EVA}: the data loader reads our fixed frame list; the KNN module is replaced by an equivalent implementation; the Pillow version is pinned; the rasterizer is recompiled; a chumpy compatibility patch is applied. Model, losses and schedule are unchanged. \emph{ExAvatar}: fitting runs in two passes (identity on the 140 fitting frames, then frozen; pose on the evaluation frames); temporal smoothing is skipped; the test background is black; the official test-pose optimization (\texttt{--fit\_pose\_to\_test}) is not used; two compatibility fixes. \emph{GaussianAvatar}: adapted from SMPL to SMPL-X; stage 1 only; background changed from white to black; checkpoint interval set to 40 so that epoch 200 is saved; two device-compatibility lines. \emph{SplattingAvatar}: the data loader reads SMPL-X meshes; official configuration; black test background; compiled with Eigen 3.4.90 instead of 3.3.7. \emph{GUAVA and OrdiHand-GS}: unchanged.

\begin{table*}[t]\centering\small
\caption{Full metrics of the comparison in the main text on the 127 annotated frames of the 60-clip subset. $\dagger$: SMPL-X adapted, stage~1. $\ddagger$: SMPL-X mesh. Hand metrics average the left- and right-hand crops. No intervals are reported.}\label{tab:supp_methods}
\begin{tabular}{lrrrrrrrrr}\toprule
&\multicolumn{3}{c}{Full frame}&\multicolumn{3}{c}{Interaction crop}&\multicolumn{3}{c}{Hand crops}\\
Method&PSNR $\uparrow$&SSIM $\uparrow$&LPIPS $\downarrow$&PSNR $\uparrow$&SSIM $\uparrow$&LPIPS $\downarrow$&PSNR $\uparrow$&SSIM $\uparrow$&LPIPS $\downarrow$\\\midrule
EVA&23.84&\textbf{0.8863}&\textbf{0.0940}&16.51&0.6942&0.3237&15.94&0.7071&0.3467\\
ExAvatar&22.40&0.8814&0.1145&16.08&0.6873&0.3510&15.21&0.6908&0.3793\\
GaussianAvatar$^\dagger$&22.34&0.8703&0.1058&15.26&0.6686&0.3435&14.78&0.6857&0.3641\\
SplattingAvatar$^\ddagger$&\textbf{24.56}&0.8837&0.1252&17.35&0.6825&0.3627&16.75&0.6963&0.3822\\
GUAVA&23.51&0.8799&0.0986&16.95&0.7064&0.3190&16.23&0.7135&0.3454\\
OrdiHand-GS (ours)&23.64&0.8807&0.0977&\textbf{17.60}&\textbf{0.7209}&\textbf{0.3026}&\textbf{16.99}&\textbf{0.7318}&\textbf{0.3236}\\\bottomrule
\end{tabular}
\end{table*}
Table~\ref{tab:supp_methods} lists all nine metrics. On the full frame the per-subject methods lead: SplattingAvatar has the highest PSNR and EVA the best SSIM and LPIPS. OrdiHand-GS has the best value on every interaction and hand metric; in a paired comparison over the 127 frames its interaction and hand PSNR differences to SplattingAvatar are not significant, whereas its remaining differences to every method are.

\noindent\textbf{Paired differences.}
Table~\ref{tab:supp_paired} reports OrdiHand-GS minus each compared method on the same 127 frames, paired by frame, with 95\% clip-clustered bootstrap intervals. On the interaction and hand crops, OrdiHand-GS is significantly better than every compared method on all three metrics, with one exception: its PSNR gains over SplattingAvatar ($+0.25$ and $+0.24$\,dB) are not significant. On the full frame, SplattingAvatar is significantly better in PSNR and SSIM and EVA in SSIM; the PSNR and LPIPS differences to EVA and the SSIM difference to ExAvatar are not significant, and OrdiHand-GS has significantly lower full-frame LPIPS than ExAvatar, GaussianAvatar, SplattingAvatar and GUAVA.

\begin{table*}[t]\centering\small
\caption{Paired differences (OrdiHand-GS minus method) on the 127 annotated frames of the 60-clip subset with 95\% clip-clustered bootstrap intervals. $^\circ$: interval includes zero. $^*$: the other method is significantly better. Unmarked: OrdiHand-GS is significantly better. Signs follow each metric's direction (positive PSNR/SSIM and negative LPIPS differences favor OrdiHand-GS).}\label{tab:supp_paired}
\setlength{\tabcolsep}{3pt}
\resizebox{\textwidth}{!}{%
\begin{tabular}{lrrrrrrrrr}\toprule
&\multicolumn{3}{c}{Full frame}&\multicolumn{3}{c}{Interaction crop}&\multicolumn{3}{c}{Hand crops}\\\cmidrule(lr){2-4}\cmidrule(lr){5-7}\cmidrule(lr){8-10}
Method&$\Delta$PSNR&$\Delta$SSIM&$\Delta$LPIPS&$\Delta$PSNR&$\Delta$SSIM&$\Delta$LPIPS&$\Delta$PSNR&$\Delta$SSIM&$\Delta$LPIPS\\\midrule
EVA&$-0.21^\circ$&$-0.0056^*$&$+0.0037^\circ$&$+1.09$&$+0.0267$&$-0.0211$&$+1.05$&$+0.0247$&$-0.0231$\\
&{\tiny[$-0.45$, $+0.41$]}&{\tiny[$-0.0081$, $-0.0000$]}&{\tiny[$-0.0007$, $+0.0054$]}&{\tiny[$+0.79$, $+1.84$]}&{\tiny[$+0.0168$, $+0.0478$]}&{\tiny[$-0.0414$, $-0.0131$]}&{\tiny[$+0.77$, $+1.77$]}&{\tiny[$+0.0145$, $+0.0449$]}&{\tiny[$-0.0427$, $-0.0151$]}\\[2pt]
ExAvatar&$+1.23$&$-0.0007^\circ$&$-0.0168$&$+1.51$&$+0.0335$&$-0.0484$&$+1.78$&$+0.0410$&$-0.0557$\\
&{\tiny[$+0.85$, $+1.85$]}&{\tiny[$-0.0039$, $+0.0034$]}&{\tiny[$-0.0203$, $-0.0144$]}&{\tiny[$+1.28$, $+1.93$]}&{\tiny[$+0.0263$, $+0.0424$]}&{\tiny[$-0.0567$, $-0.0427$]}&{\tiny[$+1.55$, $+2.16$]}&{\tiny[$+0.0336$, $+0.0498$]}&{\tiny[$-0.0640$, $-0.0495$]}\\[2pt]
GaussianAvatar$^\dagger$&$+1.29$&$+0.0104$&$-0.0081$&$+2.33$&$+0.0523$&$-0.0409$&$+2.21$&$+0.0462$&$-0.0405$\\
&{\tiny[$+1.12$, $+1.56$]}&{\tiny[$+0.0081$, $+0.0129$]}&{\tiny[$-0.0102$, $-0.0060$]}&{\tiny[$+2.10$, $+2.74$]}&{\tiny[$+0.0452$, $+0.0617$]}&{\tiny[$-0.0487$, $-0.0358$]}&{\tiny[$+1.95$, $+2.59$]}&{\tiny[$+0.0388$, $+0.0548$]}&{\tiny[$-0.0484$, $-0.0348$]}\\[2pt]
SplattingAvatar$^\ddagger$&$-0.93^*$&$-0.0031^*$&$-0.0275$&$+0.25^\circ$&$+0.0384$&$-0.0601$&$+0.24^\circ$&$+0.0355$&$-0.0586$\\
&{\tiny[$-1.15$, $-0.74$]}&{\tiny[$-0.0059$, $-0.0013$]}&{\tiny[$-0.0294$, $-0.0255$]}&{\tiny[$-0.00$, $+0.58$]}&{\tiny[$+0.0299$, $+0.0461$]}&{\tiny[$-0.0672$, $-0.0537$]}&{\tiny[$-0.02$, $+0.54$]}&{\tiny[$+0.0272$, $+0.0427$]}&{\tiny[$-0.0663$, $-0.0517$]}\\[2pt]
GUAVA&$+0.13$&$+0.0007$&$-0.0009$&$+0.65$&$+0.0144$&$-0.0164$&$+0.76$&$+0.0183$&$-0.0218$\\
&{\tiny[$+0.07$, $+0.19$]}&{\tiny[$+0.0003$, $+0.0012$]}&{\tiny[$-0.0012$, $-0.0006$]}&{\tiny[$+0.54$, $+0.81$]}&{\tiny[$+0.0110$, $+0.0174$]}&{\tiny[$-0.0198$, $-0.0138$]}&{\tiny[$+0.64$, $+0.94$]}&{\tiny[$+0.0143$, $+0.0224$]}&{\tiny[$-0.0262$, $-0.0183$]}\\\bottomrule
\end{tabular}}
\end{table*}

\noindent\textbf{Per-signer macro average.}
Because the 60 clips are unevenly distributed over the six signers (6, 4, 13, 16, 14 and 7 clips), Table~\ref{tab:supp_signer} averages each metric first within signer and then across signers. The ranking of the six methods on these three metrics is the same as in Table~\ref{tab:supp_methods}, and OrdiHand-GS remains the best on each.

\begin{table}[t]\centering\footnotesize\setlength{\tabcolsep}{4pt}
\caption{Signer-macro-averaged metrics on the 60-clip subset (mean over six signers of per-signer means).}\label{tab:supp_signer}
\resizebox{\linewidth}{!}{%
\begin{tabular}{lrrr}\toprule
Method&Inter.\ LPIPS $\downarrow$&Inter.\ PSNR $\uparrow$&Hand LPIPS $\downarrow$\\\midrule
EVA&0.3360&16.42&0.3570\\
ExAvatar&0.3539&16.30&0.3815\\
GaussianAvatar$^\dagger$&0.3477&15.49&0.3660\\
SplattingAvatar$^\ddagger$&0.3685&17.41&0.3862\\
GUAVA&0.3259&17.06&0.3517\\
OrdiHand-GS (ours)&\textbf{0.3098}&\textbf{17.73}&\textbf{0.3297}\\\bottomrule
\end{tabular}}
\end{table}

\begin{figure}[t]\centering
\includegraphics[width=\linewidth]{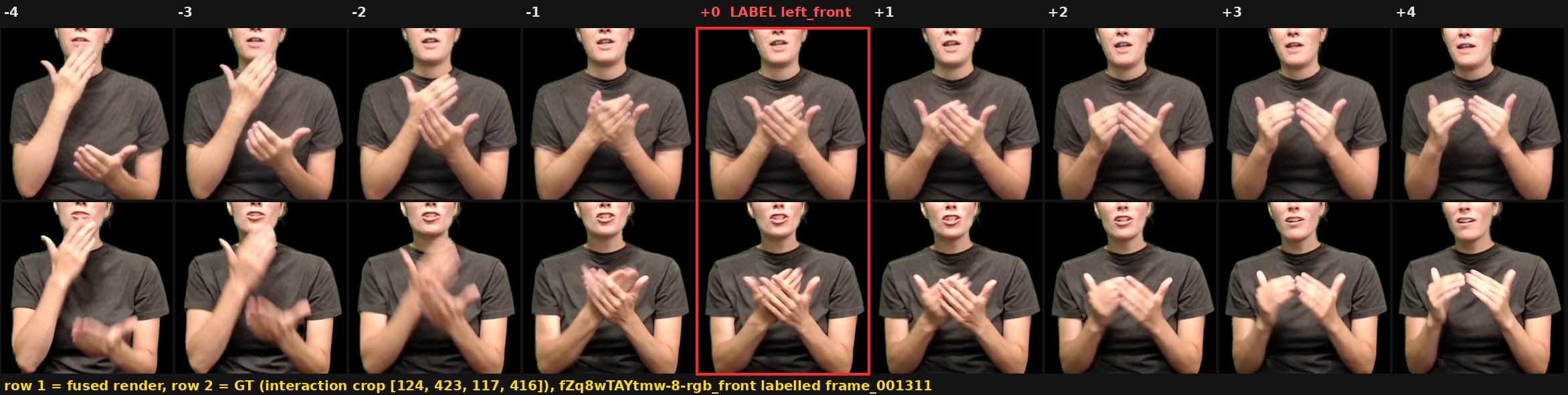}\\[2pt]
\includegraphics[width=\linewidth]{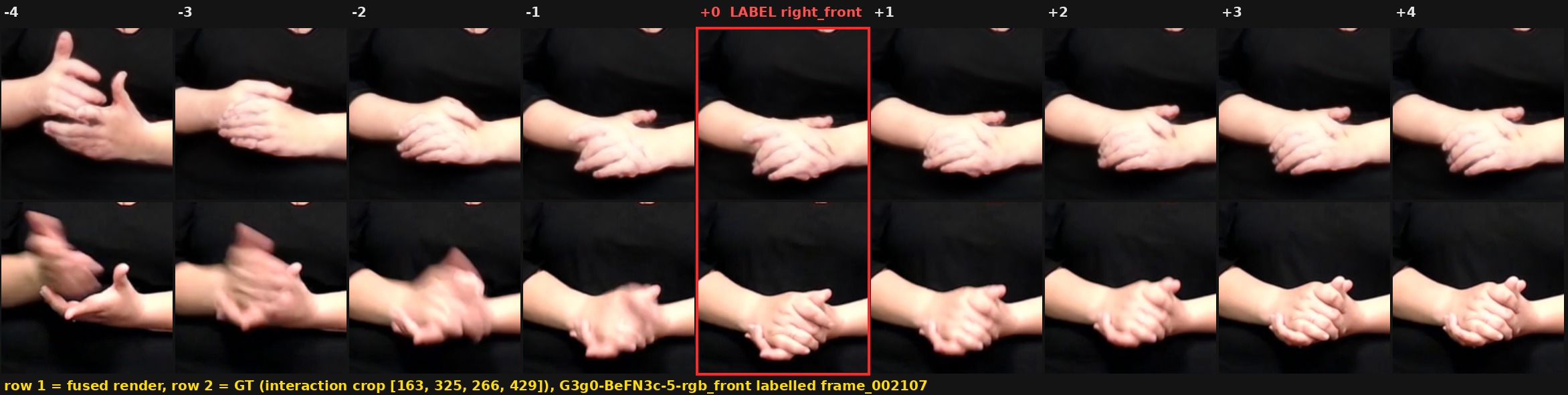}
\caption{Window-level rendering inspection. Top: w03 around the corrected frame 1311. Bottom: w06 around frame 2107, where fused tracking reverses the order incorrectly.}\label{fig:windows}
\end{figure}

\begin{figure*}[t]\centering
\includegraphics[width=0.94\textwidth]{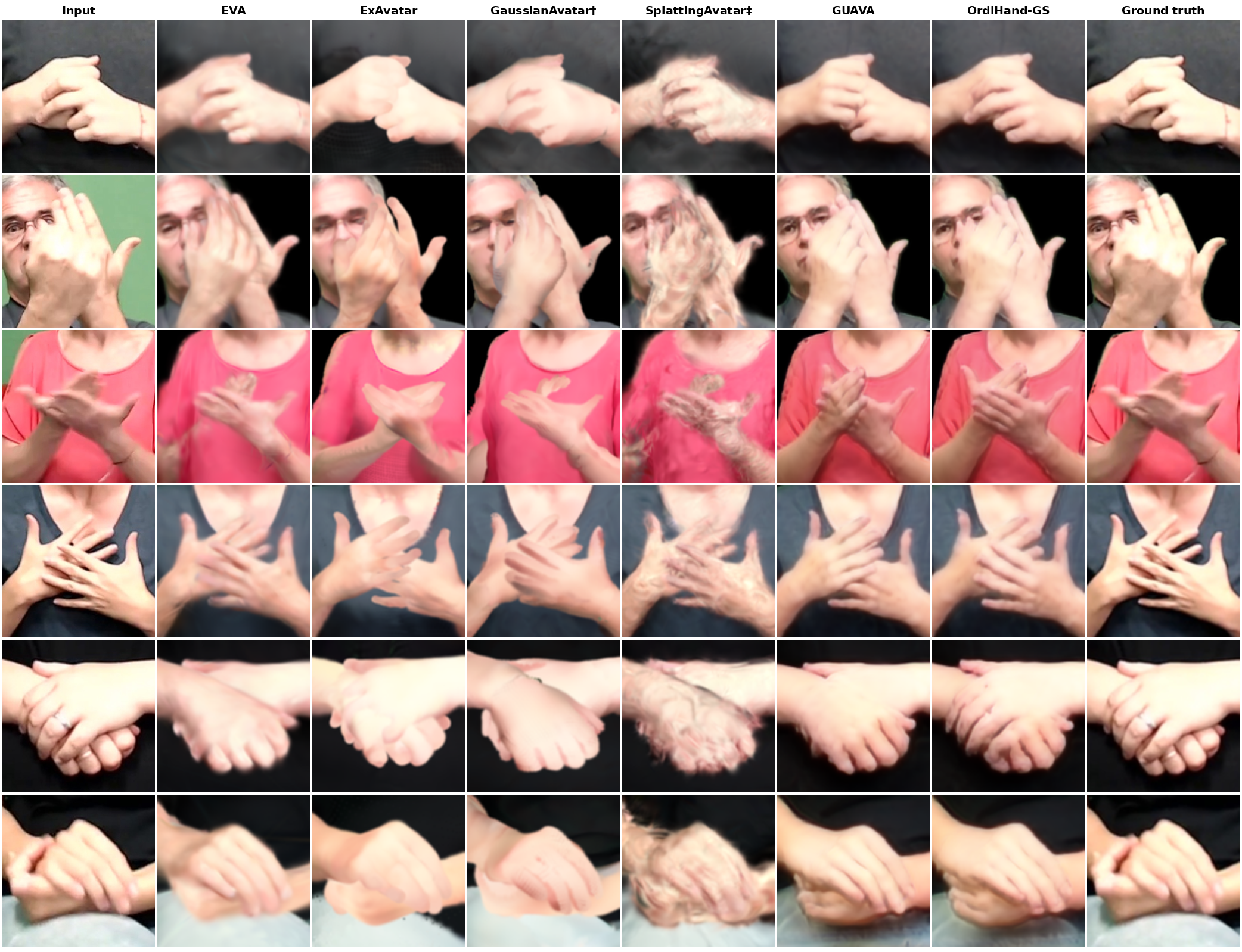}
\caption{Qualitative comparison on the 60-clip subset, one frame per signer. Columns: input, EVA, ExAvatar, GaussianAvatar$^\dagger$, SplattingAvatar$^\ddagger$, GUAVA, OrdiHand-GS, ground truth. All panels show the interaction crop of the $512\times512$ evaluation render on a black background (the input column is the unmasked video frame). Per-subject methods are fitted to 140 frames of the clip; GUAVA-based columns use one reference image.}\label{fig:supp_methods}
\end{figure*}

\noindent\textbf{Qualitative comparison.}
Figure~\ref{fig:supp_methods} shows interaction crops of all six methods on six frames of the 60-clip subset, one per signer, selected by a fixed hash of the clip and frame names that was declared before any cross-method score was computed.

\section{Window-Level Inspection}
\label{sec:supp_windows}
Figure~\ref{fig:windows} shows two of the eight slow-motion windows inspected in the main text: w03, where the cues cancel on frame 1310 and the next frame is corrected, and w06, where fused tracking reverses the order incorrectly.

\section{Hyperparameters}
\label{sec:supp_hparams}
Table~\ref{tab:supp_hparams} collects every fixed constant of the method in one place; all were set before the corresponding arm was evaluated and none was tuned on results.
\begin{table}[t]\centering\footnotesize\setlength{\tabcolsep}{3pt}
\caption{Fixed constants of OrdiHand-GS. Values are those used in all reported runs.}\label{tab:supp_hparams}
\begin{tabular}{p{4.3cm}p{3.5cm}}\toprule
Quantity & Value\\\midrule
\multicolumn{2}{l}{\emph{Frame pool and supports}}\\
Hull-overlap threshold for the annotation pool & 0.30\\
Hull dilation / intersection dilation & 8 px / 4 px\\
Minimum support size for the depth cue & 50 px\\\midrule
\multicolumn{2}{l}{\emph{Ordinal cues}}\\
Depth model & Video Depth Anything, ViT-L\\
Depth abstention threshold $|\delta_t|$ & 0.02\\
Depth saturation $\kappa_d$ & 0.10\\
Size abstention threshold $|r_t|$ & 0.01\\
Size saturation $\kappa_a$ & 0.2\\
Temporal median window (ablation only) & 5 frames ($\pm2$), median of non-abstaining votes\\\midrule
\multicolumn{2}{l}{\emph{Hinge and refinement}}\\
Margin $m$ (EHM units) & 0.03\\
Ordinal weight $\lambda_{\mathrm{ord}}$ & 100\\
Refinement steps (main / small budget) & 240 / 60\\
Hand-pose learning rate & $10^{-4}$\\
Other parameter groups & baseline settings\\\midrule
\multicolumn{2}{l}{\emph{Residual head}}\\
Architecture / parameters & 3 conv layers / 221{,}571\\
Training frames / updates / seed & 100{,}002 / 40{,}000 / 10\\
Optimizer & Adam ($\beta_1=0$, $\beta_2=0.99$), batch 6\\
Learning rate & $10^{-3}\rightarrow2.5\times10^{-4}$, linear\\
Training crop & $256\times256$\\
Image loss & $\ell_1 + 0.05\,\mathrm{LPIPS}_{\mathrm{Alex}}$\\
Hand loss & $0.1\,\ell_1 + 0.005\,\mathrm{LPIPS}_{\mathrm{Alex}}$\newline $+\,0.01\,(1-\mathrm{SSIM})$\\
Hand-term weight in the objective & 0.25\\\midrule
\multicolumn{2}{l}{\emph{Evaluation}}\\
LPIPS backbone / resolution & VGG / $280\times280$\\
Bootstrap & 95\%, clip-clustered\\\bottomrule
\end{tabular}
\end{table}

\section{Additional Failure Cases}
\label{sec:supp_failures}
\begin{figure*}[t]\centering
\includegraphics[width=0.88\textwidth]{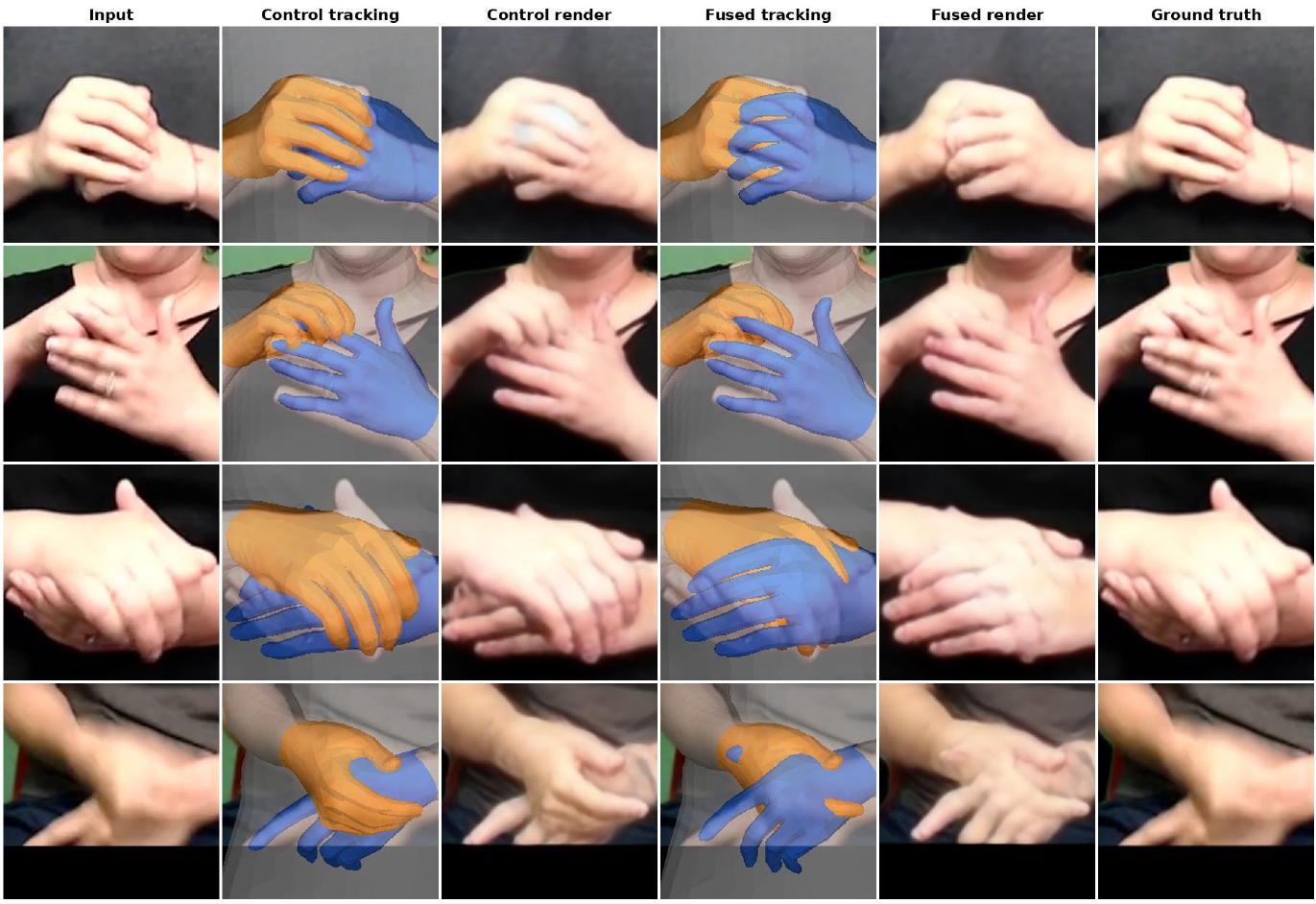}
\caption{New ordering errors introduced by fused refinement relative to the matched control (4 of 22, drawn at random). Columns: input, control tracking and its render, fused tracking and its render, ground truth; interaction crops; left hand blue, right hand orange; all renders use the original GUAVA weights.}\label{fig:supp_failures}
\end{figure*}

Relative to the matched control, fused refinement corrects 184 wrist orders and introduces 22 new errors (main text Table~1). Figure~\ref{fig:supp_failures} shows 4 of the 22 new errors, drawn at random with a fixed seed before any of these frames was inspected. In all four the annotated order is right-in-front, the control fit has the right hand in front, and fused refinement moves the left hand in front. The stored cue votes explain the direction: in the first row the depth cue votes for the left hand and outweighs a weak, correct size vote; in the second and fourth rows both cues vote for the left hand; in the third row the depth cue votes for the left hand and the size cue abstains. The hands are interlaced or stacked in the first three rows, and the input frame of the fourth row is motion-blurred. Over all 22 new errors, both cues vote for the wrong hand in 10, exactly one cue is wrong and outvotes the other in 8, one cue is wrong while the other abstains in 3, and in 1 the fused vote is correct but the optimized order is not; 21 of the 22 are annotated right-in-front, which follows from the definition: a new error requires a correct control fit, and the control fit places the right hand in front in 338 of the 365 frames.

\section{Runtime}
\label{sec:supp_runtime}
Table~\ref{tab:supp_runtime} reports the wall-clock cost of the inference-time components of OrdiHand-GS on one A800 GPU; GUAVA and OrdiHand-GS need no per-subject training (Table~\ref{tab:supp_methods_train}). VDA and refinement times are the wall-clock times recorded by the runs that produced the reported results (365 depth windows; 171 clip neighborhoods). Cue extraction was timed on the 365 annotated frames on one CPU core. Rendering was timed on the 365 annotated frames with batch size 1 after one warm-up frame per clip, and covers animation, splatting and neural refinement; it corresponds to about 51 frames per second. The residual head runs once per reference image inside the feed-forward reconstruction (58\,ms per reference image in total), not once per frame, so it adds no per-frame cost. Among the listed components, ordinal refinement and depth estimation dominate; the cost of the original EHM tracking that they build on is not included.
\begin{table}[t]\centering\footnotesize
\caption{Inference-time cost of the components of OrdiHand-GS on one A800 GPU (cue extraction on one CPU core). The residual head runs once per reference image, not per frame.}\label{tab:supp_runtime}
\begin{tabular}{p{3.3cm}p{3.9cm}}\toprule
Component & Time\\\midrule
VDA depth inference (ViT-L) & 3.3\,s per 32-frame window (0.10\,s per frame)\\
Cue extraction (hulls, votes) & 5.0\,ms per frame (CPU)\\
Ordinal EHM refinement (240 steps, per clip neighborhood) & 25.0\,s per neighborhood (1.30\,s per frame)\\
Residual-head forward pass & 2.0\,ms per reference image\\
GUAVA rendering at $512\times512$ & 19.6\,ms per frame\\\bottomrule
\end{tabular}
\end{table}

\section{Per-Signer Wrist Order}
\label{sec:supp_signers}
Table~\ref{tab:supp_signer_order} breaks the wrist-order results of main text Table~1 down by signer. Fused refinement is more accurate than the matched control for every signer, with accuracies between 67.3\% and 89.7\%. The low original and control accuracies of signer~1 reflect the label distribution of that subset (37 of 39 frames are left-in-front, while the original fit places the right hand in front in most frames).
\begin{table}[t]\centering\small
\caption{Wrist-order accuracy (\%) per signer on the 365 annotated frames. No intervals are reported for these subsets.}\label{tab:supp_signer_order}
\begin{tabular}{lrrrr}\toprule
Signer & Frames & Original EHM & Control & Fused\\\midrule
1&39&10.3&7.7&89.7\\
2&18&38.9&33.3&77.8\\
3&79&30.4&30.4&77.2\\
5&102&38.2&39.2&81.4\\
8&78&37.2&35.9&71.8\\
10&49&38.8&38.8&67.3\\\midrule
All&365&33.4&32.9&77.3\\\bottomrule
\end{tabular}
\end{table}

\end{document}